\RequirePackage[bookmarksnumbered,unicode]{hyperref}
\documentclass[sigconf,nonacm]{acmart}

\usepackage{booktabs,array,tabularx}
\newcolumntype{L}[1]{>{\raggedright\arraybackslash}p{#1}}
\newcolumntype{Y}{>{\raggedright\arraybackslash}X}

\usepackage{tikz}
\usetikzlibrary{arrows.meta,positioning,fit,backgrounds}

\begin{document}

\title{RAG Deserves an Index: Why Ingest-Time Compilation Beats
Query-Time Interpretation}

\author{Kyle Wild}
\orcid{0009-0001-6918-3197}
\affiliation{%
  \institution{Endgame Labs, Inc.}
  \city{San Francisco}
  \country{USA}}
\affiliation{%
  \institution{Asia AI Institute, Musashino University}
  \city{Tokyo}
  \country{Japan}}
\email{orcid.org/0009-0001-6918-3197}

\author{Yusuke Takahashi}
\orcid{0009-0006-8351-2280}
\affiliation{%
  \institution{Faculty of Data Science, Musashino University}
  \institution{Asia AI Institute}
  \institution{AIx, Inc.}
  \city{Tokyo}
  \country{Japan}}
\email{orcid.org/0009-0006-8351-2280}

\author{Asako Uraki}
\orcid{0009-0006-8412-1804}
\affiliation{%
  \institution{Faculty of Data Science, Musashino University}
  \institution{Asia AI Institute}
  \institution{AIx, Inc.}
  \city{Tokyo}
  \country{Japan}}
\email{orcid.org/0009-0006-8412-1804}

\begin{abstract}
Nearly every retrieval-augmented question-answering system in
production ships with a hidden interpreter: on each query a language
model re-derives the meaning of raw corpus text --- resolving
references, attributing claims, reconstructing structure --- and then
throws that work away. The problem is pernicious precisely because
frontier models do this well and keep getting better at it, and
waiting for cheaper ones does not close the gap: per-token prices have
fallen by orders of magnitude while aggregate inference spend has
risen, because context volume grows faster than prices
fall~\cite{hamilton2026tokentax}. This is the modern equivalent of the
full-table scan, and the remedy is the one databases found fifty years
ago: do the expensive work once, at write time, into a maintained
structure that makes reads cheaper. In databases that structure was
the index; a corpus whose read pattern is known before it ever meets a
user can and should be indexed too.

We call the paradigm \emph{ingest-time semantic compilation} (ISC):
compile a corpus's meaning into a queryable substrate with two coupled
layers --- incrementally maintained embeddings, and atomic claims
whose provenance is validated at compile time --- and treat that
substrate as a first-class database object with its own DDL,
maintenance contract, migration contract, and cost model. Two
existence proofs support it. Substrate upkeep scales with change
rather than corpus size: incremental updates run $33.7\times$ cheaper
than reconstruction while tracking it to floating-point precision. And
on a held-out sample of 500 broadcast-interview transcripts, compiled
claims as the retrieval payload win every one of 32 budget-by-model
cells --- 85.2\% correct from roughly 2.2k reader tokens against 72.5\%
from roughly 16.3k for the best chunk configuration anywhere. The only
baseline that keeps pace on accuracy is a contextualized-chunk
pipeline with hybrid retrieval and reranking, statistically
indistinguishable from compiled claims at roughly twenty-one times the
query-path tokens --- and it reaches that parity, we argue, precisely
because it has itself begun to compile. We close with the systems
agenda this opens, from compilation planners to read planning.
\end{abstract}



\maketitle

\section{The Interpreter We Ship on Every Query}

A modern RAG or agentic system~\cite{lewis2020rag} answers a question by retrieving raw
text and asking a large language model to make sense of it --- every
single time. If the corpus is a year of meetings and the question is
``who committed to the migration deadline?'', the model must find the
speaker, resolve ``he'' across turns, distinguish an assertion from a
quotation, and locate the one load-bearing sentence inside kilobytes of
adjacent conversation. Then the next query pays for all of it again.

Databases named this pattern long ago. An interpreter re-translates the
program on every run; a compiler translates once and reuses the result.
A system without indexes scans the table on every query, whereas an
index lets you pay the cost once, at write time, so reads are cheap
forever. Query-time semantic reconstruction --- QSR, as we will call
the dominant paradigm --- is semantically an unindexed scan performed
by the most expensive scan operator ever deployed.

The analogy is tighter than a rhetorical flourish. A database
administrator adds an index once the production read pattern is
established: the queries are known, so the work they repeat can be moved
to write time. Retrieval-augmented question answering is in that
position before it ever meets a user. We already know the shape of the
read pattern --- people will ask questions answerable by reasoning over
the semantic information living within the corpus --- and we know which
interpretive steps every such question will re-pay for. That is
precisely the condition under which a database is indexed rather than
scanned.

The costs are no longer hypothetical, and they compound.

\textbf{Economic.} Agentic workloads inflate read-time context
multiplicatively. Industry token prices per unit fell orders of
magnitude while aggregate inference spend rose, precisely because
context volume grew faster than prices
fell~\cite{hamilton2026tokentax}. Paying a model to re-derive the same
meaning on every query is the dominant marginal cost of deployed
systems.

\textbf{Behavioral.} Context is not accuracy-neutral. Model reliability
degrades measurably as inputs grow, relevant material is lost in the
middle of long contexts, and irrelevant material actively
distracts~\cite{liu2024lost,hong2025contextrot,shi2023distracted}. More
retrieval is not more accuracy; often it is less. Our own measurements
below reproduce this directly.

\textbf{Epistemic.} Nothing in the QSR pipeline distinguishes what a
source actually said from what the model inferred while reading it.
Every query is a fresh opportunity to silently re-guess meaning, and
memory-system benchmarks now measure exactly this failure propagating
from extraction into answers~\cite{chen2025halumem}.

The database community has seen this movie. So the question this
paper asks is this: what would it mean to build the compiler?

\section{The Vision: A Compiled Semantic Substrate}
\label{sec:vision}

Ingest-time semantic compilation performs the semantic work when a
document arrives and persists it as a \emph{semantic substrate} the
system consults thereafter. We propose the substrate as a first-class
database object with two coupled layers.

\textbf{The geometric layer} --- embeddings and their index
structures, maintained incrementally as the corpus and the embedding
models evolve. This layer answers \emph{where is meaning like this?}

\textbf{The symbolic layer} --- atomic, self-contained claims compiled
from source text, each carrying non-negotiable provenance: the verbatim
source span, its author, and its location, validated mechanically at
compile time. This layer answers \emph{what exactly was said, by whom?}
--- and, critically, it is the payload the reading model consumes, not
merely an index key that points back into raw text. That distinction
--- compiling the payload, not just the pointer --- separates ISC from
contextualized chunking and proposition indexing, which improve where
retrieval lands while the reader still consumes raw
text~\cite{chen2024densex,anthropic2024contextual}.

Concretely, instead of returning a 1,000-token transcript window in
which a deadline is discussed somewhere, the substrate returns a claim
that already carries its own evidence:

\begin{quote}\small
Priya Shah committed to deliver the migration plan by September~30.\\
\emph{Evidence:} ``I'll have the migration plan ready by the end of
September.''\\
\emph{Source:} Priya Shah, turn 42, project-planning meeting.
\end{quote}

\noindent
The reader receives the claim, the span that supports it, and the
attribution, and has to recover none of the three. The original
transcript remains available for inspection, verification, and the
cases where the compiled form has dropped context the question needs.

Around these layers, the paradigm imports the full apparatus databases
attach to any maintained derived structure.

\textbf{A compilation contract (semantic DDL).} What gets compiled,
from what sources, under what validation gate. A claim that fails
exact-quote validation does not enter the substrate: provenance behaves
like a foreign-key constraint on meaning.

\textbf{A maintenance contract.} Sources change; substrates must track
them at a cost proportional to change, not corpus size --- the property
that makes materialized views viable, demanded of semantics.

\textbf{A migration contract.} Embedding models are the substrate's
storage format, and they churn. Model upgrades must be absorbable as
alignment problems, not full rebuilds.

\textbf{A cost model.} With ingest cost per document, maintenance cost
per change, and per-query savings, the break-even read count $R^*$ is
an optimizer statistic: compile when expected reads exceed it,
interpret when they do not. Compilation is a spectrum, and where a
corpus sits on it should be a planned decision, not an architectural
accident.

\begin{table}[t]
\centering
\caption{The four contracts, what each guarantees, the mechanism that
enforces it, and where this paper exhibits it.}
\label{tab:contracts}
{\scriptsize
\setlength{\tabcolsep}{3.2pt}
\renewcommand{\arraystretch}{1.15}
\begin{tabularx}{\columnwidth}{@{}YYYc@{}}
\toprule
\textbf{Contract} & \textbf{Guarantees} & \textbf{Enforced by} &
\textbf{Shown} \\
\midrule
Compilation (semantic DDL) & only source-supported claims enter &
exact-quote validation gate & \S2.1 \\
Maintenance & cost tracks change, not corpus size & incremental
low-rank update & \S3.1 \\
Migration & a model upgrade is an alignment problem & orthogonal
Procrustes & \S3.1 \\
Cost model & compile only where reads repay it & $R^*$ as a planner
statistic & App.~A \\
\bottomrule
\end{tabularx}
}
\end{table}

\subsection{What a compiled claim actually is}
\label{sec:claim}

Concretely, in our reference implementation the substrate is ordinary
PostgreSQL. A compiled claim is a row in \texttt{facts} --- the
self-contained claim text, an optional topic and
subject--relation--target triple, a content hash, a lifecycle state,
and a \texttt{pipeline\_run\_id} recording exactly which extractor,
model, and prompt produced it --- joined to one or more rows in
\texttt{fact\_evidence}: the verbatim quote, its character offsets, its
turn range, and the method by which the quote was located. Provenance
is not an annotation on this schema; it is its foreign keys. A fact
references its source document, its evidence references the fact, and
\texttt{UNIQUE(document\_id, content\_hash)} makes recompilation
idempotent.

The validation gate sits between extraction and insertion. A candidate
claim must carry a supporting quote; the quote must be located in the
canonical transcript, and the located span must resolve to real turns
and a real speaker. A candidate that fails --- typically a
model-``quoted'' sentence that is actually a paraphrase, so the string
search finds no span --- is not repaired or approximated. It is dropped
before it touches the substrate, exactly as a row violating a
foreign-key constraint never enters a table. On the held-out corpus the
gate admitted 69,746 claims, every one carrying a supporting quote
located byte-exactly in its source transcript; in a 20-document replay
of the frozen extraction, the gate rejected 1.1\% of candidate claims
(29 of 2,724), 28 of them for quotes that could not be located in the
source --- the class of unsupported compilation that extraction-based
memory systems are measured to leak~\cite{chen2025halumem}.

Maintenance rides on the same rows: \texttt{state} and
\texttt{superseded\_by} implement tombstoning and revision without
destroying lineage, and an \texttt{index\_outbox} table propagates
upserts and deletes to the vector index --- the substrate's redo log in
miniature, whose production form Section~\ref{sec:agenda} asks the
community to design.

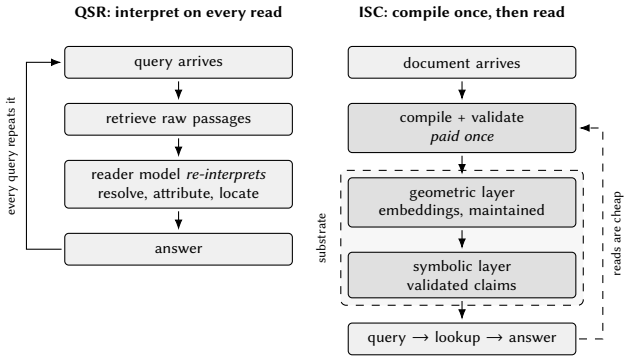
\begin{figure}[t]
\centering
\begin{tikzpicture}[
  font=\sffamily\scriptsize,
box/.style={draw, rounded corners=1.5pt, align=center,
              inner sep=3pt, minimum height=4.2mm, text width=28mm},  raw/.style={box, fill=black!6},
  cmp/.style={box, fill=black!12},
  lbl/.style={font=\sffamily\bfseries\scriptsize},
  a/.style={-{Latex[length=1.4mm]}, shorten >=1pt, shorten <=1pt},
  node distance=3.2mm]

\node[lbl] (qh) at (0,0) {QSR: interpret on every read};
\node[raw, below=2.2mm of qh] (q1) {query arrives};
\node[raw, below=of q1] (q2) {retrieve raw passages};
\node[raw, below=of q2] (q3) {reader model \emph{re-interprets}\\resolve, attribute, locate};
\node[raw, below=of q3] (q4) {answer};
\draw[a] (q1)--(q2); \draw[a] (q2)--(q3); \draw[a] (q3)--(q4);
\draw[a] (q4.west) -- ++(-5mm,0) |- node[pos=0.25,rotate=90,anchor=south,
  font=\sffamily\tiny] {every query repeats it} (q1.west);
  
\node[lbl] (ih) at (3.75,0) {ISC: compile once, then read};
\node[raw, below=2.2mm of ih] (i1) {document arrives};
\node[cmp, below=of i1] (i2) {compile + validate\\\emph{paid once}};
\node[cmp, below=of i2] (i3) {geometric layer\\embeddings, maintained};
\node[cmp, below=of i3] (i4) {symbolic layer\\validated claims};
\node[raw, below=of i4] (i5) {query $\to$ lookup $\to$ answer};
\draw[a] (i1)--(i2); \draw[a] (i2)--(i3); \draw[a] (i3)--(i4); \draw[a] (i4)--(i5);

\begin{scope}[on background layer]
  \node[draw, dashed, rounded corners=2pt, fill=black!3,
        fit=(i3)(i4), inner sep=2.6pt] (sub) {};
\end{scope}
\node[font=\sffamily\tiny, left=0.8mm of sub, rotate=90, anchor=south]
  {substrate};
  
\draw[a, dashed] (i5.east) -- ++(3.5mm,0) |- node[pos=0.25,rotate=90,
  anchor=north, font=\sffamily\tiny] {reads are cheap} (i2.east);
  \end{tikzpicture}
\Description{Two parallel flow diagrams. On the left, query-time
semantic reconstruction: a query arrives, raw passages are retrieved,
a reader model re-interprets them, and an answer is produced, with
the cycle repeating on every query. On the right, ingest-time
semantic compilation: a document arrives and is compiled and
validated once into a substrate of maintained embeddings and
validated claims, after which a query becomes a cheap lookup.}
\caption{Query-time semantic reconstruction repeats the interpretive
work on every read; ingest-time semantic compilation performs it once
and maintains the result.}
\label{fig:substrate}
\end{figure}

\section{Two Existence Proofs}

\subsection{Maintenance scales with change, not size}
\label{sec:maint}

In a controlled synthetic pilot on an evolving substrate --- a corpus
of synthetic embedding vectors grown from 3,000 to 9,000 documents over
50 update events, constructed to isolate maintenance cost from corpus
confounds --- incremental low-rank updates cost 8.4~ms per update
against 283~ms for full re-decomposition, $33.7\times$ cheaper per
update and $23.8\times$ cheaper cumulatively, while tracking the fully
recomputed subspace to floating-point precision: maximum
principal-angle drift below $10^{-11}$ degrees, recall@10 of
1.0~\cite{takahashi2026substrate}. The shape matters as much as the
ratio: incremental per-update cost stayed flat as the corpus tripled,
while full re-decomposition rose with $N$, so the gap widens rather
than closes as a corpus grows. The cheap path is not a lossy
shortcut. A feared embedding-model migration was absorbed by an
orthogonal Procrustes virtual-axis update that recovered 0.95 mean
cosine to truly re-embedded vectors while re-embedding only about 10\%
of the corpus.

The standard objection to compiled semantic structures --- ``you can't
afford to keep them current'' --- is, on this evidence, misplaced: a
large, slowly-changing corpus is precisely where maintenance is cheap
and QSR re-pays the most. The pilot is synthetic and its incremental
update idealized, so these are best-case bounds; validation on real
corpora under production embedding APIs is the study's stated next
step.

\subsection{Compiled payloads dominate the read-time frontier}
\label{sec:readpath}

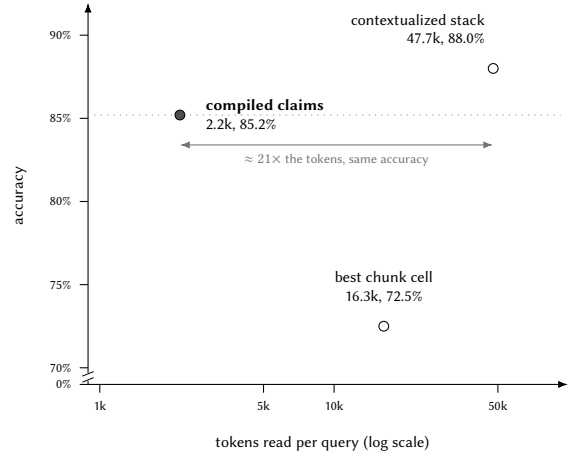
\begin{figure}[t]
\centering
\begin{tikzpicture}[font=\sffamily\scriptsize,
  pt/.style={circle, draw, fill=black!70, inner sep=1.3pt},
  ptw/.style={circle, draw, fill=white, inner sep=1.3pt}]
\def\X#1{{(#1-2.95)*3.10}}
\def\Y#1{{(#1-69)*0.22}}
\draw[-{Latex[length=1.3mm]}] (0,0) -- (6.35,0);
\node[font=\sffamily\scriptsize] at (3.1,-0.78)
  {tokens read per query (log scale)};
\draw[-{Latex[length=1.3mm]}] (0,0) -- (0,5.05);
\node[rotate=90, anchor=south, font=\sffamily\scriptsize] at (-0.70,2.5)
  {accuracy};
\draw[white, line width=1.6pt] (0,0.045) -- (0,0.155);
\draw[line width=0.35pt] (-0.075,0.03) -- (0.075,0.09);
\draw[line width=0.35pt] (-0.075,0.11) -- (0.075,0.17);
\foreach \l/\t in {3/1k, 3.6989/5k, 4/10k, 4.6989/50k}{
  \draw (\X{\l},0)--(\X{\l},-1.1mm) node[below, font=\sffamily\tiny] {\t};}
\draw (0,0)--(-1.1mm,0) node[left, font=\sffamily\tiny] {0\%};
\foreach \a in {70,75,80,85,90}{
  \draw (0,\Y{\a})--(-1.1mm,\Y{\a}) node[left, font=\sffamily\tiny] {\a\%};}
\draw[dotted, black!45] (0,\Y{85.2}) -- (6.25,\Y{85.2});
\node[pt] (f) at (\X{3.342},\Y{85.2}) {};
\node[ptw] (c) at (\X{4.212},\Y{72.5}) {};
\node[ptw] (r) at (\X{4.679},\Y{88.0}) {};
\node[anchor=west, font=\sffamily\scriptsize, align=left]
  at ([xshift=1.6mm]f.east) {\textbf{compiled claims}\\2.2k, 85.2\%};
\node[anchor=south, font=\sffamily\scriptsize, align=center]
  at ([yshift=1.2mm]c.north) {best chunk cell\\16.3k, 72.5\%};
\node[anchor=south east, font=\sffamily\scriptsize, align=right]
  at ([yshift=1.2mm]r.north) {contextualized stack\\47.7k, 88.0\%};
\draw[{Latex[length=1.3mm]}-{Latex[length=1.3mm]}, black!55]
  (\X{3.342},\Y{83.4}) -- node[below, font=\sffamily\tiny, black!55]
  {$\approx21\times$ the tokens, same accuracy} (\X{4.679},\Y{83.4});
\end{tikzpicture}
\Description{A scatter plot of accuracy against tokens read per
query on a log scale. Compiled claims sit at 2.2k tokens and 85.2
percent accuracy in the upper left. The best chunk cell sits at 16.3k
tokens and 72.5 percent, below and to the right. The contextualized
stack sits at 47.7k tokens and 88.0 percent, far to the right at
roughly the same height as compiled claims.}
\caption{The read-time frontier on the held-out sample. Compiled
claims occupy the upper left; the only configuration that matches
them on accuracy sits far to the right, and every configuration that
reads raw text sits below. The vertical axis is truncated below
69\% (break marked); exact values are printed beside each point.}
\label{fig:frontier}
\end{figure}

In a controlled dialogue question-answering study on broadcast-interview
transcripts~\cite{zhu2021mediasum}, provenance-validated facts serving
as the retrieval payload were compared against three chunking policies
--- fixed-width, turn-aware, and semantic --- under identical
embedding, retrieval, and answer models across matched read-token
budgets.

Results are reported on a held-out sample of 500 transcripts and 499
questions sharing no document with the sample used to develop
extraction.\footnote{Three runs of one chunk baseline recorded recall
of 0.000 because retrieval executed before the embedding index was
built. They were deleted before any judging and re-run with the
embedding step in place; no judged result depends on them.} Facts won
every one of 32 budget-by-model cells, and all 24
facts-versus-chunk comparisons survived Holm correction for multiple
testing ($p \le 4.9\times10^{-6}$; strongest cell
$1.4\times10^{-42}$) under McNemar's test on paired per-question
outcomes. At a 2,048-token budget facts answered 85.2\% correctly from
roughly 2.2k reader tokens, while the best chunk configuration anywhere
in the sweep reached 72.5\% from roughly 16.3k. Facts at a 256-token
budget (69.5\%) beat every chunk policy at 2,048.
Figure~\ref{fig:frontier} places the three headline configurations on
the accuracy-per-token plane.

A decomposition into coverage and reading accuracy explains the shape.
Fact payloads carried the gold evidence in 98--99\% of cases and stayed
flat as the budget grew, while chunk reading accuracy declined from
81\% to 73\% as more text was supplied. The advantage is a
model-general reading-quality gap over raw transcript, not retrieval
placement --- and supplying more raw context is not free even when it
contains the answer.

Against the strongest chunk-side remedy --- LLM-contextualized chunks
with hybrid dense and full-text retrieval fused by reciprocal rank,
plus a cross-encoder reranker, pre-registered before execution --- the
two approaches are statistically indistinguishable. At a 2,048-token
budget the full stack answered 88.0\% (439/499) against 85.2\%
(425/499) for facts, McNemar $p = 0.202$; at 16,384 the figures are
87.4\% and 83.6\%, $p = 0.076$. Neither difference survives correction
for multiple testing. The nominal direction reversed between
development and held-out: on the frozen development sample facts led
92 to 88 ($p = 0.39$), while on held-out the full stack leads by 2.8 to
3.8 points. We report the reversal rather than the more convenient
half of it, because the finding is the tie and not its sign --- and
because the tie is bought at roughly $21\times$ the query-path tokens,
about 47.7k against 2.2k, on every query forever.

The stack's own components do not reach that tie. Its vector and
hybrid arms answer 73.9\% and 75.6\%, losing to facts at Holm-corrected
$p \le 1.4\times10^{-4}$. Only the assembled stack keeps pace, and its
mechanism analysis is the interesting part: it works precisely insofar
as it, too, performs compilation --- the situating preamble names
referents a reader would otherwise resolve --- while still lacking
validated provenance, since nothing checks the generated preamble
against the source span. The strongest available chunk pipeline is an
incomplete instance of the architecture we propose, paying at query
time for work a compiler could validate once.

That pattern is this paper's claim in miniature. Every configuration
that interprets at query time loses decisively; the only one that keeps
pace has already begun to compile, and pays twenty-one times over for
the half of the job it does at read time. Full protocol and results
appear in a companion report~\cite{wild2026iscrag}.

Compilation is not free, and we can price it. Extraction runs Kimi
K2.6 turn-wise over each transcript. On a 20-document replay of the
frozen extraction it averaged 48.2k prompt and 4.5k completion tokens
per document, about \$0.064 per document at current hosted pricing;
extrapolated across the corpus that is roughly \$32 and 26.3M tokens to
compile all 500 documents. The replayed sample yields about 3.6\% fewer
claims per document than the corpus average, so the extrapolation is a
slight underestimate. Set the figure against the contextualized stack's
extra query-path load: its $\sim$45.5k additional tokens per query mean
that compiling the entire corpus costs what roughly 580 of its queries
cost, and even if the full 3.6\% carried through to cost that number
moves only to about 600. For this corpus, compilation amortizes inside
the first thousand reads either way.\footnote{The $\sim$580 figure is a
token-count equivalence rather than a price one: compile-time and
query-time tokens are billed on different models. The dollar version
depends on reranker pricing, which varies by deployment.} Answer
grading in both studies is automated, and the grader is also Kimi K2.6,
so extraction and judgment share a model; human calibration and a
re-judgment under a different model family are in progress. Until they
land we treat the read-path result as strong preliminary evidence
rather than a settled effectiveness claim.

It follows that compiling is sometimes the wrong call, and the cost
model says when. A corpus that is small, highly volatile, or rarely
queried may never reach $R^*$: the ingest bill is paid in full while
the per-read saving accrues too slowly to recover it, and per-query
reconstruction remains the right architecture. We are not arguing that
meaning should always be compiled. We are arguing that whether to
compile it is a planning decision with a computable answer, and that
today it is usually made by default rather than by estimate.

One existence proof is synthetic and geometric, the other real dialogue
and symbolic. Together they bound the paradigm from both sides: the
substrate can be kept current for less than reconstruction, and reading
from it beats reading raw text at a fraction of the cost.

\section{The Research Agenda}
\label{sec:agenda}

ISC opens a systems agenda with the shape and texture of classical
database research.

\textbf{1. The compilation planner.} Not everything deserves
compilation. $R^*$ varies per document with query popularity,
volatility, and compilation cost; the planner's job is choosing,
continuously, what to compile, to what depth --- embeddings only?
facts? enriched relations? --- and what to leave interpreted. This is
the optimizer problem restated for meaning, including its adversary:
LazyGraphRAG-style arguments~\cite{msr2024lazygraphrag} that deferral
wins at low read volume are the interpreter half of the trade-off, and
the field needs the cost model that decides per corpus rather than a
doctrinal winner.

\textbf{2. Provenance as an integrity constraint.} The symbolic
layer's exact-quote gate is a constraint system: every compiled claim
is verifiable against its source span by string comparison, making
hallucinated compilation mechanically rejectable at write time --- the
property extraction-based memory systems measurably lack. What is the
full constraint language? Span containment, speaker attribution,
temporal validity, revision lineage; violations as compile errors, not
runtime surprises.

\textbf{3. Maintenance and migration at production scale.}
Incremental subspace tracking and Procrustes alignment worked in a
controlled pilot. The production questions are streaming variants under
real revision loads, anchor selection for migration, bounded staleness
contracts, and the semantic analogue of crash recovery: what is the
substrate's redo log?

\textbf{4. The substrate as shared infrastructure.} A compiled
substrate is naturally multi-tenant: many agents, assistants, and
applications reading one maintained semantic layer through standard
interfaces. This raises classical questions in new clothes --- isolation
(whose compilation contract?), versioning (queries against yesterday's
meaning), and access control at the granularity of claims.

\textbf{5. Reading as a planned operation.} Once payloads are
compiled, ``how much context to read'' becomes a per-query
physical-plan choice with measurable accuracy and cost curves ---
including the non-monotonicity we observed directly, where more context
reduced accuracy. Read planning against a substrate, with the reader
model's degradation curve as a cost-model input, is unexplored
territory.

\textbf{6. How to evaluate a compiler.} An agenda that asks for
planners needs a way to tell a good one from a bad one, and end-to-end
answer accuracy is too coarse: it conflates what was compiled, what was
retrieved, and what was read. The substrate's own properties are
separately measurable --- extraction coverage against the source,
provenance validity, self-containment, duplication, staleness under a
revision stream, and the rejection rate of the validation gate --- and
a planner should be judged on the decisions it makes given those, not
only on the answers that eventually come out. Establishing that
measurement layer is prerequisite to comparing compilation strategies
at all.

\textbf{7. Honest boundaries.} Our dialogue evidence also contains a
caution: when a tool-using agent mediated retrieval, the
compiled-payload advantage did not automatically survive the agent
layer. The interaction between compiled substrates and agentic read
strategies --- when the agent should ask for siblings, lineage, or raw
source --- is open, and we flag it rather than paper over it.

\section{Related Visions}

Document expansion by query
prediction~\cite{nogueira2019doc2query} is the oldest instance of the
instinct we follow: it spends compute at ingest, generating queries a
document could answer and appending them before indexing, so retrieval
lands better later. What it compiles is the pointer --- the expansion
changes what the index matches, while the reader still receives the
original passage --- and nothing validates the generated queries
against the source. The lineage matters: moving work to ingest time in
order to improve retrieval is not new. What is new here is compiling
the payload the reader consumes, and validating it against the span it
came from. Proposition-level indexing~\cite{chen2024densex} decomposes
passages into standalone propositions and indexes those, improving
where retrieval lands; the reader still receives raw passage text, so the
pointer is compiled and the payload is not. Contextualized
chunking~\cite{anthropic2024contextual} prefixes each chunk with a
model-generated situating preamble, a partial compilation of exactly
the reference-resolution work we target --- but nothing validates the
preamble against the source span, and the payload again remains raw
text. Agent-memory
systems~\cite{chhikara2025mem0,rasmussen2025zep} compile payloads,
which is the right unit, but without a validation gate: benchmarks show
extraction errors propagating into downstream
answers~\cite{chen2025halumem}. GraphRAG-family systems compile
structure for multi-hop reasoning, trading ingest cost for traversal
ability, and address a different query class than ours.
Context-compression methods shrink reads lossily at query time, which
is interpretation made cheaper rather than compilation. And
cost-performance analyses of fact-based memory against long context
measure the frontier we care about without asking what guarantees the
facts carry~\cite{pollertlam2026beyond}. Each holds one piece.
Table~\ref{tab:positioning} lays the field out along the two axes that
matter here. Several systems compile something; only one validates
what it compiled against the source it came from.

\begin{table}[t]
\centering
\caption{What each approach compiles, whether the compiled form is
validated against its source, and what the reading model actually
receives. Only the last row fills both of the middle columns.}
\label{tab:positioning}
{\scriptsize
\setlength{\tabcolsep}{3.2pt}
\renewcommand{\arraystretch}{1.1}
\begin{tabularx}{\columnwidth}{@{}YYcY@{}}
\toprule
\textbf{Approach} & \textbf{Compiles} & \textbf{Val.} &
\textbf{Reader receives} \\
\midrule
Document expansion~\cite{nogueira2019doc2query} & the pointer & no &
raw passage plus queries \\
Proposition indexing~\cite{chen2024densex} & the pointer & no &
raw passage \\
Contextual chunking~\cite{anthropic2024contextual} & pointer, part of
payload & no & raw passage plus preamble \\
Agent memory~\cite{chhikara2025mem0,rasmussen2025zep} & the payload &
no & extracted claims \\
GraphRAG family & structure & no & summaries, traversals \\
Context compression & nothing; shrinks at read & --- & lossy passage \\
\midrule
\textbf{ISC} & \textbf{the payload} & \textbf{yes} & \textbf{claim with
its source span} \\
\bottomrule
\end{tabularx}
}
\end{table}

The database lineage --- materialized views, inverted indexes, latent
semantic indexing --- supplies the missing discipline: derived
structures earn their keep through maintenance contracts, integrity
constraints, and cost models. ISC is the proposal that meaning deserves
the same treatment.

\section{Conclusion}

Fifty years of data systems can be read as one lesson: move work from
read time to write time, then govern the derived structure with
diligence and precision. Language models made meaning computable but
left it interpreted. Ingest-time semantic compilation is the claim that
it is now compilable --- affordably maintained, mechanically validated,
and cheaper to read than the raw text it was compiled from. Two
controlled studies support feasibility. The agenda --- planners,
constraints, migration, shared substrates, read planning --- is wide
open, and it looks like database research because it is.

\begin{acks}
Part of this work used computational infrastructure provided by
Endgame Labs, Inc. and AIx, Inc.
\end{acks}

\appendix
\renewcommand{\theHsection}{Appendix.\Alph{section}}
\section{The Break-Even Read Count}

Section~\ref{sec:vision} claims $R^*$ is an optimizer statistic. Here
it is.

Let a corpus hold $N$ documents and receive $W$ changes over the period
of interest. Write $c_c$ for the ingest cost of compiling one document,
$c_m$ for the maintenance cost of absorbing one change, $c_q$ for the
per-query cost of query-time semantic reconstruction, and $c_r$ for the
per-query cost of reading the compiled substrate. Over $R$ reads the two
architectures cost
\[
  \text{QSR}: R\,c_q
  \qquad\qquad
  \text{ISC}: N c_c + W c_m + R\,c_r ,
\]
and setting them equal gives the break-even read count
\[
  R^* = \frac{N c_c + W c_m}{c_q - c_r} .
\]

Three properties make this a planner statistic rather than an accounting
identity. First, $R^*$ is finite only when $c_q > c_r$: the compiled
path must actually be cheaper to read, which is what
Section~\ref{sec:readpath} measures. Second, the maintenance term
$W c_m$ grows with change rather than with $N$, which is what
Section~\ref{sec:maint} measures --- so a large but stable corpus does
not inflate $R^*$ merely by being large. Third, $R^*$ is a per-document
quantity in practice: $N$ collapses to 1 when the planner asks whether
to compile a particular document, and the decision turns on that
document's expected read count against its own compilation cost. A
planner that estimates read demand per document, rather than per corpus,
is the object we are asking the community to build.

The held-out corpus instantiates the numerator directly. With $N = 500$,
a $c_c$ of roughly 48.2k prompt and 4.5k completion tokens per document
measured on a 20-document replay and extrapolated, and no corpus change
during the study ($W = 0$), compilation cost about 26.3M tokens in
total. The denominator is the per-read gap:
the contextualized-chunk stack, the only baseline that keeps pace on
accuracy, spends roughly 45.5k more query-path tokens per question than
a substrate read. Dividing gives $R^* \approx 580$ reads --- for a
500-document corpus, compilation pays for itself inside the first
thousand questions. The equivalence is in tokens rather than dollars,
since compile-time and query-time work is billed on different models,
but the shape of the answer does not depend on that: $R^*$ here is a
number a planner could act on, not an asymptotic argument.

One term is idle in this instance. The study held its corpus fixed, so
$W = 0$ and the maintenance term vanishes; a deployed substrate over a
live corpus would carry it. That is exactly the term
Section~\ref{sec:maint} bounds, and the reason the two studies belong
in one paper: the read-path study fixes the denominator, the
maintenance study keeps the numerator from growing with the corpus,
and neither alone would make $R^*$ a quantity a planner could
estimate.


\end{document}